%% file: main.tex
\documentclass{article}
\usepackage{iclr2027_conference,times}
\iclrfinalcopy 
\input{math_commands.tex}

\usepackage{hyperref}
\usepackage{url}
\usepackage{graphicx}
\usepackage{booktabs}
\usepackage{amsmath}
\usepackage{amssymb}
\usepackage{multirow}
\usepackage{xcolor}
\usepackage{colortbl}
\usepackage{enumitem}
\usepackage{tikz}
\usetikzlibrary{arrows.meta,positioning,fit,backgrounds,calc}
\graphicspath{{figures/}}
\definecolor{oursA}{RGB}{226,239,247}
\newcommand{\method}{StateWise}
\newcommand{\ci}[2]{{\scriptsize[#1, #2]}}

\newcommand{\arm}[1]{\textsc{#1}}

\title{StateWise: Teaching Language Models to Track and Obey Evolving Constraints in Multi-Turn Dialogue}

\author{Samarth Varshney
\And
Ritu Yadav}

\begin{document}
\maketitle
\lhead{Preprint}  
\input{sec_abstract}
\input{sec_intro}
\input{sec_related}
\input{sec_problem}
\input{sec_method}
\input{sec_setup}
\input{sec_results}
\input{sec_analysis}
\input{sec_conclusion}
\input{sec_statements}
\bibliographystyle{iclr2027_conference}
\bibliography{references}
\appendix
\input{sec_appendix}
\end{document}

%% file: math_commands.tex
\usepackage{amsmath,amsfonts,bm}

\def\eqref#1{equation~\ref{#1}}

\def\1{\bm{1}}

\DeclareMathAlphabet{\mathsfit}{\encodingdefault}{\sfdefault}{m}{sl}
\SetMathAlphabet{\mathsfit}{bold}{\encodingdefault}{\sfdefault}{bx}{n}



%% file: sec_abstract.tex
\begin{abstract}
Users of conversational assistants rarely state all requirements up front: they add constraints, amend them, withdraw them, digress, and return. Instruction-tuned large language models (LLMs) degrade in this regime in two ways---active constraints stop being honored as turns accumulate (persistence failure), and withdrawn constraints keep being enacted (relapse)---and the only existing repair we are aware of is an external ledger that re-injects a compiled specification at every turn. To tackle this, we propose \method{}, a self-improving approach that teaches a single LLM to maintain its own revisable constraint state. We formalize evolving-constraint dialogue as edit trajectories over a typed state with tombstones for revoked constraints, so that persistence and relapse are measured by executable checkers and ground-truth states are compiled for free. \method{} first fine-tunes the LLM on two complementary skills, summarizing a dialogue into its constraint state and responding conditioned on that state; it then self-trains on trajectories the model itself generates, keeping only turns that pass an exact round-trip check against the compiled state. This recipe teaches 7--8B models to track the state almost perfectly but, surprisingly, not to obey it: on Llama-3.1-8B every supervised variant roughly doubles relapse, and on Qwen2.5-7B none reduces it. We therefore add a preference stage on self-generated relapse contrasts, in which the rejected response enacts a revoked rule and the chosen response passes the round-trip check. Experiments on two backbones with held-out constraint families and unseen phrasings show that \method{} cuts relapse by 8--22 points relative to supervised self-training, lifts conversation success by 8--19 points over the backbone, and beats an oracle compiled-ledger baseline on relapse with no inference-time scaffolding. We also characterize the recipe's cost on cumulative-constraint and single-turn benchmarks and study weight interpolation as a mitigation.
\end{abstract}

%% file: sec_intro.tex
\section{Introduction}

Conversational assistants are asked to follow instructions that evolve. A user asks for a summary in bullet points, then wants it under a hundred words, then drops the bullets, digresses to an unrelated question, and returns to the summary expecting the word limit to still hold and the bullets to be gone. Multi-turn instruction following is now benchmarked at scale: Multi-IF \citep{he2024multiif} extends IFEval \citep{zhou2023ifeval} to three turns and finds accuracy falling with every turn; SysBench \citep{qin2024sysbench} measures the instability of system-message adherence; EvolIF \citep{jia2026evolif} constructs evolving dialogues with constraint changes and reports deficiencies in failure recovery. A parallel line isolates a specific pathology: when a user withdraws a constraint, the model often keeps enacting it, a failure that ReBIND \citep{zhu2026rebind} names \emph{revocation inertia} and shows rising from 0.01 to 0.40 for an 8B model as the constraint load grows.

Training-side work has addressed adjacent problems. Reinforcement learning with verifiable rewards improves single-turn multi-constraint following \citep{pyatkin2025ifbench}; label-free variants derive rewards from the instructions themselves \citep{ren2026selfsup}; curriculum RL with abstention rewards mitigates the lost-in-conversation gap for information revealed across turns \citep{laban2025lost,li2026rlaar}. None of these trains the model to hold a \emph{revisable} constraint state across edits. To our knowledge the only published repair for relapse is inference-time: ReBIND keeps an external ledger with tombstones and compiles the net specification into every request, at roughly $1.5\times$ delivery cost and only for code.

We propose \method{}, a self-improving approach that teaches a single LLM to maintain its own revisable constraint state. Our design follows a recipe proven effective for label-free domain adaptation: SimRAG \citep{xu2025simrag} fine-tunes one LLM on complementary skills, uses the same model to synthesize training data, and keeps only examples that survive a round-trip consistency check. \method{} transposes this recipe to evolving-constraint dialogue. The complementary skills are constraint-state summarization and state-conditioned response generation; the unlabeled resource is single-turn verifiable constraints, which we compose into scripted \textsc{add}/\textsc{amend}/\textsc{revoke} trajectories; and the round-trip check is exact equality between the model's emitted state and the compiled ground truth, together with executable checkers and a relapse guard. Because the state is compiled from the edit history, every training signal and metric in this paper is verifiable without human annotation.

The recipe teaches the state and, surprisingly, fails to teach obedience to it. On Llama-3.1-8B-Instruct the self-improved model emits the correct state, including the correct tombstones, for 92\% of in-distribution turns, and then enacts the revoked rule anyway at more than twice the backbone's rate. On Llama the same doubling of relapse appears in every supervised variant we trained, with or without state supervision, and even single-turn constraint training raises relapse: making a model better at obeying constraints makes withdrawn constraints stickier. Rejection-sampling SFT removes relapsed examples from the training set but supplies no signal \emph{against} relapse. We therefore add a second self-improvement stage that turns the round-trip check into a preference signal: the model's own relapsed generations become rejected responses, its round-trip-clean generations become chosen responses, and a short DPO run on these self-labeled contrasts cuts relapse by 8--22 points, to below the backbone and below an oracle compiled-ledger system, on two backbones, two training seeds and unseen phrasings. Ablations show that the relapse-specific contrast is necessary (preference pairs about active constraints alone \emph{raise} relapse) and that label-shuffled pairs do nothing, so the gain is the signal, not the extra training.

Our contributions are summarized as follows:
\begin{itemize}[leftmargin=*,itemsep=1pt,topsep=2pt]
\item We formalize evolving-constraint dialogue as edit trajectories over a typed constraint state with tombstones, with executable metrics for persistence and relapse (the latter restricted to constraints whose chance-satisfaction rate is measured to be low), and show with a preregistered inference-only diagnostic that an 8B model fails on both and that prompting it to write its own state makes it worse.
\item We propose \method{}, a self-improving two-stage recipe with a verifiable round-trip filter and a relapse-contrastive preference stage, requiring no proprietary model and no human dialogue annotation, and we report the negative intermediate result that motivates the preference stage: tracking a state and obeying it are separable skills, and self-training with a filter teaches only the first.
\item Experiments on two backbones, held-out constraint families, unseen phrasings and public benchmarks, all under gates fixed before the runs, show relapse reductions of 8--22 points relative to supervised self-training and conversation-success gains of 8--19 points over the backbone; we also report the recipe's cost on cumulative-constraint and single-turn benchmarks and show that weight interpolation with the backbone removes that cost at the price of the relapse gain, exposing a controllable trade-off.
\end{itemize}

%% file: sec_related.tex
\section{Related Work}

\paragraph{Multi-turn instruction following.}
Multi-IF \citep{he2024multiif}, SysBench \citep{qin2024sysbench}, EvolIF \citep{jia2026evolif} and MultiTurnInstruct \citep{wang2025multiturninstruct} document degradation with dialogue depth, topic switches and revised goals; conversational-reliability studies \citep{laban2025lost} show that sharding information over turns lowers accuracy. These are evaluation resources. We use Multi-IF as a held-out test and add what it lacks: constraints that are amended and withdrawn, with the relapse they cause measured directly.

\paragraph{Verifiable instruction following.}
IFEval \citep{zhou2023ifeval} introduced programmatic checkers; IFBench \citep{pyatkin2025ifbench} added held-out constraint families and showed that RLVR generalizes to them; \citet{ren2026selfsup} removed external supervision with constraint decomposition and constraint-wise pseudo-labels. All train on single-turn prompts. \method{} uses the same style of checkers but trains on multi-turn edit trajectories with explicit state targets; single-turn constraint training is one of our baselines, and we find that it raises persistence and relapse together.

\paragraph{Lost in conversation and dialogue state.}
RLAAR \citep{li2026rlaar} adds abstention rewards and a curriculum for information revealed across turns; rolling-memory policies compress fragmented history \citep{rollingmemory2026}; history-cleaned self-distillation removes self-contamination \citep{historyclean2026}. These address information fragmentation, not constraint revision. Classic dialogue state tracking predicts slot values over a fixed ontology for task-oriented systems \citep{budzianowski2018multiwoz}; VLK-RL \citep{vlkrl2026} verifies LLM-elicited constraints for a separate RL policy. \method{}'s state is an open set of executable constraint predicates with parameters and tombstones, supervised by self-generated data, and lives in the same model that responds.

\paragraph{Revocation inertia.}
ReBIND \citep{zhu2026rebind} measures per-clause influence in code-generation dialogues and repairs relapse with an external compiled ledger via the API alone. We adopt its tombstone semantics for ground truth, train the model rather than wrap it, extend from code to general formatting constraints, and compare against ReBIND-style compiled ledgers as baselines, including a negative-priming variant that names the withdrawn rules.

\paragraph{Self-improvement with round-trip filtering and preference signals.}
SimRAG \citep{xu2025simrag} is the methodological template: complementary skills, self-generated data and a consistency filter. STaR-style rejection sampling \citep{zelikman2022star} and self-instruct \citep{wang2023selfinstruct} are the broader family; self-rewarding and iterative DPO \citep{yuan2024selfrewarding,rafailov2023dpo} turn a model's own judgments into preference data. Our filter is exact match on a structured state plus executable checks, not retrieval or LLM judgment, and our preference pairs are labeled by the same checkers, so no judge model is involved in the pipeline.

%% file: sec_problem.tex
\section{Problem Formulation}
\label{sec:problem}

\paragraph{Constraints, edits and state.}
A constraint is a pair $c=(f,\theta)$ of a family $f$ and parameters $\theta$ with an executable checker $\mathrm{check}_f(y,\theta)\in\{0,1\}$ over a response $y$; examples are \texttt{all\_caps}, \texttt{max\_words}(80), \texttt{end\_phrase}(``Is there anything else?''), \texttt{json\_only} and \texttt{forbidden\_word}(``very''). Twelve families are used in training and six further families appear only at test time (Appendix~\ref{app:families}). A dialogue applies one or more edits per user turn: \textsc{add}$(c)$, \textsc{amend}$(c\rightarrow c')$, \textsc{revoke}$(c)$, \textsc{distract} (an unrelated request) and \textsc{return} (back to the task). The net state after turn $t$ is $S_t=(A_t,R_t)$: the active set $A_t$ holds at most one constraint per family, and the tombstone set $R_t$ holds every revoked or superseded constraint. \textsc{amend} moves the old version to $R_t$; re-adding a tombstoned constraint revives it. $S_t$ is compiled from the edit list, has an order-independent canonical serialization, and is therefore exact ground truth for both training and evaluation at no annotation cost.

\paragraph{Metrics.}
\emph{Persistence} at turn $t$ is the fraction of $A_t$ satisfied by the response $y_t$. \emph{Relapse} at turn $t$ is the fraction of $R_t$ satisfied by $y_t$, restricted to families whose chance-satisfaction rate $p_0\le 0.10$, where $p_0$ is measured on the backbone's own unconstrained responses to the same task prompts (a response that happens to contain no commas is not evidence of relapse for \texttt{no\_commas} unless commas are otherwise common); an amended constraint whose new version logically implies the old one is not counted. \emph{Conversation success} requires every active constraint satisfied and zero relapse at every turn. All metrics are averaged over conversations and reported with conversation-level bootstrap intervals; comparisons between arms are paired on shared trajectory identifiers.

\paragraph{Trajectory synthesis.}
Scripted trajectories are generated from a seed: turn 0 adds one or two constraints and requests a short task; later turns \textsc{add}, \textsc{amend} or \textsc{revoke} with at most three constraints active and incompatible pairs (e.g., \texttt{json\_only} with \texttt{bullets\_exact}) never jointly active; half of the trajectories insert a \textsc{distract} turn followed by a \textsc{return}; every trajectory contains at least one \textsc{revoke} or \textsc{amend}; the final turn is always a task turn. Each user message is composed from natural-language phrasings of the edits plus a task prompt, so every assistant turn is checkable. Trajectories have four to eight turns. Three evaluation sets are used throughout: an in-distribution set (240 conversations, training families and phrasings, unseen seed), a held-out set (300 conversations; 18 families of which 6 are never trained on), and a paraphrase set (300 conversations; author-written alternative phrasings and new task prompts). Test sets are fixed before any training run.

%% file: sec_method.tex
\section{Methodology}
\label{sec:method}

\begin{figure}[t]
\centering
\resizebox{\textwidth}{!}{%
\begin{tikzpicture}[font=\sffamily\small, node distance=4mm and 5mm,
  box/.style={rounded corners=3pt, line width=0.7pt, align=center, inner sep=4pt, font=\sffamily\small, text width=36mm},
  data/.style={box, draw=black!50, fill=black!6},
  skill/.style={box, draw=blue!60!black, fill=blue!8},
  filt/.style={box, draw=orange!80!black, fill=orange!10},
  pref/.style={box, draw=purple!70!black, fill=purple!8},
  arr/.style={-{Stealth[length=2mm]}, line width=0.8pt, draw=black!60},
  stage/.style={rounded corners=5pt, draw=black!20, fill=black!2, inner sep=6pt}]
\node[data] (fam) {single-turn verifiable constraint families};
\node[data, below=of fam] (traj) {scripted edit trajectories (\textsc{add}/\textsc{amend}/\textsc{revoke}) + compiled states $S_t$};
\node[skill, below=of traj, text width=36mm] (s1) {skill 1: dialogue prefix $\rightarrow$ constraint state\\[1pt] skill 2: prefix + state $\rightarrow$ response};
\begin{scope}[on background layer]\node[stage, fit=(fam)(traj)(s1)] (p1) {};\end{scope}
\node[font=\sffamily\bfseries\small, anchor=south west] at ([yshift=1mm]p1.north west) {Stage I: skill induction};
\draw[arr] (fam) -- (traj); \draw[arr] (traj) -- (s1);
\node[data, right=9mm of fam] (gen) {model generates state $\hat S_t$ + response $y_t$ on fresh trajectories};
\node[filt, below=of gen] (rt) {round-trip filter: $\hat S_t = S_t$ exactly; all active checkers pass; no low-chance relapse};
\node[skill, below=of rt] (rft) {rejection-sampling SFT on kept turns};
\begin{scope}[on background layer]\node[stage, fit=(gen)(rt)(rft)] (p2) {};\end{scope}
\node[font=\sffamily\bfseries\small, anchor=south west] at ([yshift=1mm]p2.north west) {Stage II: self-training};
\draw[arr] (gen) -- (rt); \draw[arr] (rt) -- (rft);
\node[pref, right=9mm of gen] (pairs) {self-generated contrasts: rejected enacts a tombstoned rule; chosen passes the round-trip};
\node[pref, below=of pairs] (dpo) {relapse-contrastive DPO (frozen reference, 59 steps)};
\node[data, below=of dpo] (out) {\method{}: emits its state block, then answers; no ledger at inference};
\begin{scope}[on background layer]\node[stage, fit=(pairs)(dpo)(out)] (p3) {};\end{scope}
\node[font=\sffamily\bfseries\small, anchor=south west] at ([yshift=1mm]p3.north west) {Stage III: preference learning};
\draw[arr] (pairs) -- (dpo); \draw[arr] (dpo) -- (out);
\draw[arr] (s1.east) -- ++(4.5mm,0) |- node[pos=0.25, right, font=\sffamily\scriptsize]{} (gen.west);
\draw[arr] (rft.east) -- ++(4.5mm,0) |- (pairs.west);
\end{tikzpicture}}
\caption{Overview of \method{}. Stage I induces two complementary skills from scripted trajectories whose ground-truth states are compiled from the edit history. Stage II lets the model generate state and response on fresh trajectories and keeps only turns that pass an exact round-trip check. Stage III turns the same check into a preference signal: the model's relapsed generations are rejected responses, its round-trip-clean generations are chosen responses. No proprietary model, judge or human annotation is used in training.}
\label{fig:overview}
\end{figure}
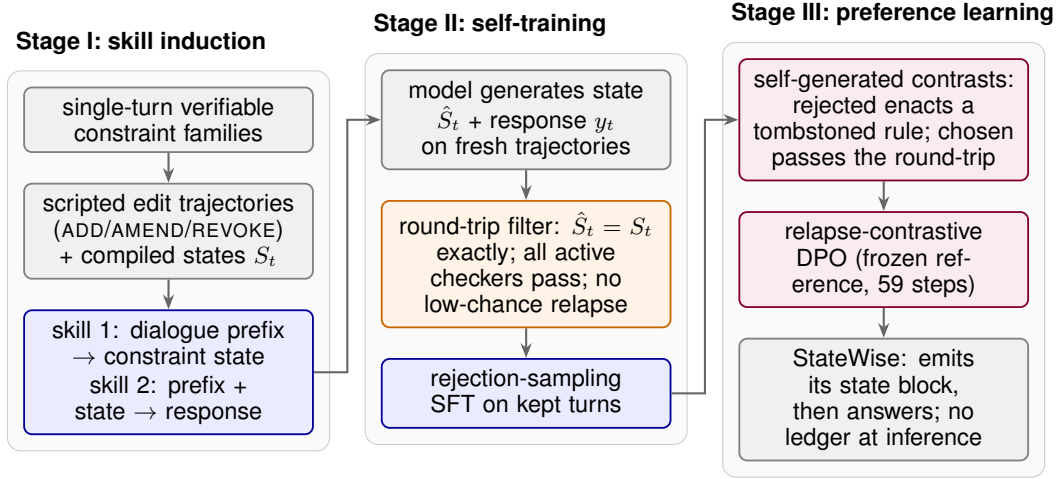

\subsection{Stage I: Skill Induction}
Starting from an instruction-tuned backbone (Figure~\ref{fig:overview}), we fine-tune with LoRA on three mixtures built from scripted trajectories (Section~\ref{sec:problem}): (a) single-turn verifiable instruction following, as in IFEval; (b) \emph{state summarization}, where the input is a dialogue prefix and the target is the canonical JSON of the compiled state $S_t$, listing active constraints with parameters and revoked constraints as tombstones; and (c) \emph{state-conditioned generation}, where the input is the prefix plus the ground-truth state block and the target is a response that passes every active checker, obtained by rejection sampling from the backbone. Mixture sizes and hyper-parameters were fixed before training (Appendix~\ref{app:train}). Only the dialogue and the constraint library are needed; the backbone is its own teacher.

\subsection{Stage II: Self-Training with Round-Trip Filtering}
The Stage-I model generates, for each turn of fresh trajectories, a state block followed by a response. A turn is kept if and only if (i) the emitted state equals the compiled state exactly, (ii) every active checker passes, and (iii) no low-chance tombstoned constraint is satisfied. Per-family shares are capped to avoid collapse onto easy families. The model is fine-tuned on the kept turns (rejection-sampling SFT). Filter yield and rejection reasons are in Appendix~\ref{app:filter}; on both backbones relapse is the dominant rejection reason, foreshadowing Section~\ref{sec:results}: the filter removes relapsed examples but supplies no signal against relapse.

\subsection{Stage III: Relapse-Contrastive Preference Learning}
The Stage-II model is sampled on a further set of trajectories (a different seed from Stage II). For each turn we pair a \emph{rejected} response that enacts a tombstoned rule with a \emph{chosen} response from the same prefix that passes the full round-trip check; a capped share (30\%) of pairs contrasts an active-constraint violation with a passing response instead. The pairs are labeled entirely by the checkers and the compiled state. We run DPO \citep{rafailov2023dpo} with a frozen reference (the Stage-II weights with adapters disabled), $\beta=0.1$, an auxiliary NLL weight of 0.5 on the chosen response, and a fixed budget of 59 optimizer steps over 931 pairs (947 constructed, 663 relapse and 284 active-violation, of which 16 were dropped for length). Nothing is tuned after seeing results.

\subsection{Inference}
In \emph{state mode} the model emits its state block and then answers; the block is stripped before the response is checked. In \emph{state-free mode} the same weights are prompted without the state instruction, which tests whether the behavior is internalized. We also evaluate \emph{weight interpolation} $\theta_\alpha=\theta_{\text{base}}+\alpha(\theta_{\text{Stage III}}-\theta_{\text{base}})$ for $\alpha\in\{0.5,0.75\}$, with no training, as a way to trade the relapse gain against breadth on general instruction following (Section~\ref{sec:breadth}).

%% file: sec_setup.tex
\section{Experimental Setup}
\label{sec:setup}

\paragraph{Backbones and arms.}
Llama-3.1-8B-Instruct is the primary backbone (two training seeds) and Qwen2.5-7B-Instruct the secondary (one seed, full pipeline with Qwen as its own teacher). The arms are: \arm{A0} the backbone; \arm{A1} single-turn constraint training (rejection-sampling SFT on single-turn verifiable prompts, the IFBench-style recipe); \arm{A2} Stage I only; \arm{A3} Stages I+II without state supervision (same trajectories, checker-only filter); \arm{A4} Stages I+II with state (the self-training recipe); \arm{A5} \arm{A4} in state-free mode; \arm{A9} Stage I with an active-only state block (no tombstones); \arm{A7} \arm{A4} plus Stage III (\method{}); and two ReBIND-style external-ledger baselines around the untrained backbone at matched prompt tokens, \arm{A6}, which compiles the net specification including the withdrawn rules under a ``do not follow'' header, and \arm{A6b}, which lists only the rules in force. \arm{A6b} uses the compiled ground-truth state at inference and is therefore an \emph{oracle} ledger.

\paragraph{Evaluation sets and metrics.}
Held-out (300 conversations, 18 families, 6 unseen), in-distribution (240 conversations) and paraphrase (300 conversations) sets (Section~\ref{sec:problem}), all with greedy decoding. Public benchmarks: Multi-IF \citep[English, 909 conversations, prompt-level strict accuracy at turns 1--3 with the model's own history]{he2024multiif}; IFEval \citep[541 prompts, prompt-level strict and loose]{zhou2023ifeval}; MT-Bench first-turn pairwise preference against the backbone judged by gemma-2-9b-it in both presentation orders (a weak local judge; regression signal only). Metrics are persistence, relapse and conversation success (Section~\ref{sec:problem}), state exact-match for state-mode arms, and 95\% conversation-level bootstrap intervals (2,000 resamples); deltas between arms are paired on trajectory identifiers. The backbone's own greedy results move by a few points between runs because batched bf16 non-determinism compounds across turns, so only within-run paired comparisons are used.

\paragraph{Preregistration.}
Hypotheses, arms, metrics and pass/fail gates were fixed before each run (Appendix~\ref{app:prereg}) and not changed after inspection; failed gates are reported as failures. The main gates are G1 (\arm{A4} beats \arm{A0} on success and relapse), G2 (state supervision helps: \arm{A4} vs.\ \arm{A3}), H2 (gain survives state-free inference), H3 (trained model matches the external ledger), H5 (Stage III beats \arm{A0} on success and relapse with intervals excluding zero), and, for the extension, replication of direction on a second seed and backbone, a Multi-IF direction check, a no-regression gate on IFEval (within 2 points), and two Stage-III ablation gates.

\paragraph{Training details.}
LoRA rank 64, learning rate $10^{-4}$ for SFT stages; DPO as in Section~\ref{sec:method}. Stage I uses 9.2k examples, Stage II the round-trip survivors of 2,400 self-generated trajectories (per-family caps), Stage III 931 pairs. Each training run uses two A100-40GB GPUs; on Llama-3.1-8B, Stage I takes about an hour, Stage II 68 minutes and Stage III 17 minutes. Full hyper-parameters, data counts and filter statistics are in Appendix~\ref{app:train}.

%% file: sec_results.tex
\section{Experimental Results}
\label{sec:results}

\subsection{The Failure Exists and Prompting Does Not Fix It}
\label{sec:stage0}
Before training we ran a preregistered inference-only diagnostic on 240 trajectories (1,438 assistant turns) with Llama-3.1-8B-Instruct (Table~\ref{tab:stage0}). Plain chat honors 79\% of active constraints overall, falling from 0.887 at the first turn to 0.653 at the eighth, and enacts revoked low-chance constraints on 11\% of post-edit turns. A system prompt asking the model to write its own constraint state before answering makes every metric \emph{worse}: persistence $-0.21$, relapse $+0.05$, success $-0.19$, all with intervals excluding zero, and the emitted states match the compiled ground truth 0.3\% of the time (omitted rules, family confusions, constraints hallucinated from task content, and an always-empty revoked list). The failure is real at the 8B operating point, prompting alone does not supply a usable state, and an external ledger must therefore be compiled programmatically, as ReBIND does, rather than requested from the model. This is the premise for training.

\begin{table}[t]
\centering\small
\resizebox{\textwidth}{!}{%
\begin{tabular}{lcccc}
\toprule
\textbf{Arm (no training)} & \textbf{Persistence} & \textbf{Relapse} & \textbf{Conv.\ success} & \textbf{State match} \\
\midrule
Plain chat & 0.793 \ci{0.767}{0.820} & 0.109 \ci{0.080}{0.141} & 0.275 \ci{0.225}{0.333} & -- \\
State-scratchpad prompt & 0.587 \ci{0.547}{0.629} & 0.162 \ci{0.125}{0.200} & 0.088 \ci{0.054}{0.125} & 0.003 \ci{0.001}{0.006} \\
Paired delta & $-0.206$ \ci{-0.253}{-0.160} & $+0.052$ \ci{0.014}{0.090} & $-0.188$ \ci{-0.250}{-0.125} & \\
\bottomrule
\end{tabular}}
\caption{Inference-only diagnostic (Llama-3.1-8B-Instruct, 240 conversations). Asking the untrained model to maintain its own constraint state makes it worse on every metric.}
\label{tab:stage0}
\end{table}

\subsection{Main Results}
\label{sec:main}

\begin{table}[t]
\centering\footnotesize
\setlength{\tabcolsep}{3.5pt}
\resizebox{\textwidth}{!}{%
\begin{tabular}{llcccccc}
\toprule
& & \multicolumn{3}{c}{\textbf{Held-out (300 conv., 6 unseen families)}} & \multicolumn{3}{c}{\textbf{In-distribution (240 conv.)}} \\
\cmidrule(lr){3-5}\cmidrule(lr){6-8}
& \textbf{Arm} & \textbf{Success} & \textbf{Relapse} & \textbf{Persist.} & \textbf{Success} & \textbf{Relapse} & \textbf{Persist.} \\
\midrule
\multirow{2}{*}{\rotatebox{90}{\tiny none}} & \arm{A0} backbone & 0.290 \ci{0.240}{0.343} & 0.130 \ci{0.100}{0.165} & 0.799 & 0.275 \ci{0.221}{0.333} & 0.110 \ci{0.081}{0.141} & 0.795 \\
& \arm{A1} single-turn constraint SFT & 0.280 \ci{0.230}{0.330} & 0.164 \ci{0.131}{0.197} & 0.847 & 0.329 \ci{0.271}{0.388} & 0.148 \ci{0.114}{0.183} & 0.856 \\
\midrule
\multirow{2}{*}{\rotatebox{90}{\tiny ledger}} & \arm{A6} oracle ledger incl.\ withdrawn rules & 0.203 \ci{0.160}{0.250} & 0.380 \ci{0.334}{0.424} & 0.871 & 0.242 \ci{0.188}{0.300} & 0.333 \ci{0.284}{0.383} & 0.887 \\
& \arm{A6b} oracle ledger, active rules only & 0.390 \ci{0.337}{0.443} & 0.148 \ci{0.117}{0.182} & 0.877 & 0.463 \ci{0.400}{0.525} & 0.084 \ci{0.058}{0.114} & 0.890 \\
\midrule
\multirow{5}{*}{\rotatebox{90}{\tiny self-training}} & \arm{A2} Stage I (state) & 0.250 \ci{0.203}{0.300} & 0.254 & 0.841 & 0.325 & 0.281 & 0.926 \\
& \arm{A9} Stage I, active-only state & 0.293 \ci{0.247}{0.343} & 0.276 & 0.882 & 0.350 & 0.261 & 0.916 \\
& \arm{A3} Stages I+II, no state & 0.310 \ci{0.257}{0.363} & 0.277 & 0.878 & 0.371 & 0.228 & 0.897 \\
& \arm{A4} Stages I+II, state & 0.327 \ci{0.273}{0.380} & 0.246 & 0.853 & 0.350 & 0.258 & 0.937 \\
& \arm{A5} \arm{A4}, state-free inference & 0.240 \ci{0.193}{0.290} & 0.246 & 0.825 & 0.292 & 0.203 & 0.855 \\
\midrule
& \cellcolor{oursA}\arm{A7} \method{} (\arm{A4} + Stage III) & \cellcolor{oursA}\textbf{0.370} \ci{0.313}{0.423} & \cellcolor{oursA}\textbf{0.094} \ci{0.069}{0.121} & \cellcolor{oursA}0.834 & \cellcolor{oursA}\textbf{0.554} \ci{0.492}{0.617} & \cellcolor{oursA}\textbf{0.042} \ci{0.024}{0.062} & \cellcolor{oursA}0.904 \\
\bottomrule
\end{tabular}}
\caption{Main results on Llama-3.1-8B-Instruct (training seed 1, greedy decoding). Conversation-level 95\% bootstrap intervals; \arm{A6}/\arm{A6b} receive the compiled ground-truth state at inference and are oracle baselines. State exact-match for the state arms: \arm{A2} 0.527 / 0.981 (held-out / in-dist.), \arm{A4} 0.510 / 0.916, \arm{A7} 0.500 / 0.893.}
\label{tab:main}
\end{table}

Table~\ref{tab:main} presents the preregistered arms on the primary backbone. From the results, we have the following findings:

(1) \emph{Every supervised arm learns to track the state and roughly doubles relapse.} \arm{A4} emits the exact compiled state, tombstones included, for 92\% of in-distribution turns and raises persistence by five points, yet its relapse is 0.246 held-out against the backbone's 0.130 ($+0.116$ \ci{0.078}{0.151}, paired). The same doubling appears without state supervision (\arm{A3}, 0.277) and with an active-only state (\arm{A9}, 0.276), and even single-turn constraint training raises relapse (\arm{A1}, 0.164 from 0.130): making a model better at obeying constraints makes withdrawn constraints stickier. State supervision helps only marginally (\arm{A4} vs.\ \arm{A3}: relapse $-0.031$ \ci{-0.058}{-0.004}), and the gain does not survive removing the state at inference (\arm{A5}). Under the preregistered gates, G1, H2 and H3 fail and G2 passes narrowly. Tracking the state and obeying it are separable skills, and rejection-sampling SFT teaches the first without the second: the filter removes relapsed examples but supplies no signal against relapse.

(2) \emph{The relapse-contrastive stage supplies the missing signal.} Fifty-nine DPO steps on 931 self-generated contrasts cut relapse from 0.246 to 0.094 held-out ($-0.152$ \ci{-0.186}{-0.120}) and from 0.258 to 0.042 in distribution ($-0.216$ \ci{-0.258}{-0.177}), to below the backbone ($-0.036$ \ci{-0.065}{-0.011} held-out; $-0.068$ \ci{-0.099}{-0.040} in-dist.) while keeping the persistence gain ($+0.035$ \ci{0.002}{0.067} over \arm{A0}) and lifting conversation success ($+0.080$ \ci{0.020}{0.140} held-out; $+0.279$ \ci{0.204}{0.350} in-dist.). State exact-match is unchanged (0.893 vs.\ 0.916 in-dist.): the gain is in obeying the state, not in tracking it. Gate H5 passes.

(3) \emph{\method{} matches or beats an oracle external ledger without scaffolding.} \arm{A6b}, the untrained backbone wrapped in a programmatically compiled active-only ledger with ground-truth knowledge of every edit, is the strongest baseline. \method{} is below it on relapse everywhere ($-0.054$ \ci{-0.081}{-0.030} held-out; $-0.042$ \ci{-0.070}{-0.017} in-dist.), matches it on held-out success ($-0.020$ \ci{-0.080}{0.043}) and beats it in distribution ($+0.092$ \ci{0.029}{0.158}); the remaining gap is four points of persistence. The ledger variant that names withdrawn rules under a ``do not follow'' header (\arm{A6}) nearly triples relapse (0.380): any salient mention of a revoked rule, even negated, primes the 8B model to enact it. This priming does not transfer to a trained self-emitted state: \arm{A9}, whose state block omits the tombstones, relapses no less than \arm{A2}, which names them ($+0.022$ \ci{-0.004}{0.051}).

\subsection{Replication: Second Seed, Second Backbone, Unseen Phrasings}
\label{sec:replication}

\begin{table}[t]
\centering\footnotesize
\setlength{\tabcolsep}{4pt}
\resizebox{\textwidth}{!}{%
\begin{tabular}{llcccccc}
\toprule
\textbf{Setting} & \textbf{Arm} & \textbf{Success} & \textbf{Relapse} & \textbf{Persist.} & \multicolumn{3}{c}{\textbf{Paired delta vs.\ \arm{A0} (success / relapse)}} \\
\midrule
\multirow{4}{*}{Llama-3.1-8B, seed 2, held-out} & \arm{A0} & 0.250 & 0.124 & 0.788 & \multicolumn{3}{c}{} \\
& \arm{A6b} oracle ledger & 0.430 & 0.145 & 0.884 & \multicolumn{3}{c}{} \\
& \arm{A4} self-training & 0.337 & 0.263 & 0.885 & \multicolumn{3}{c}{} \\
& \cellcolor{oursA}\arm{A7} \method{} & \cellcolor{oursA}\textbf{0.443} & \cellcolor{oursA}\textbf{0.097} & \cellcolor{oursA}0.860 & \multicolumn{3}{c}{\cellcolor{oursA}$+0.193$ \ci{0.133}{0.253} / $-0.026$ \ci{-0.056}{0.004}} \\
\midrule
\multirow{4}{*}{Qwen2.5-7B, held-out} & \arm{A0} & 0.163 & 0.230 & 0.731 & \multicolumn{3}{c}{} \\
& \arm{A6b} oracle ledger & 0.260 & 0.244 & 0.865 & \multicolumn{3}{c}{} \\
& \arm{A4} self-training & 0.277 & 0.220 & 0.830 & \multicolumn{3}{c}{} \\
& \cellcolor{oursA}\arm{A7} \method{} & \cellcolor{oursA}\textbf{0.357} & \cellcolor{oursA}\textbf{0.144} & \cellcolor{oursA}0.837 & \multicolumn{3}{c}{\cellcolor{oursA}$+0.193$ \ci{0.137}{0.253} / $-0.086$ \ci{-0.118}{-0.053}} \\
\midrule
\multirow{4}{*}{Llama-3.1-8B, paraphrase set} & \arm{A0} & 0.243 & 0.168 & 0.809 & \multicolumn{3}{c}{} \\
& \arm{A6b} oracle ledger & \textbf{0.403} & 0.163 & 0.919 & \multicolumn{3}{c}{} \\
& \arm{A4} self-training & 0.260 & 0.268 & 0.867 & \multicolumn{3}{c}{} \\
& \cellcolor{oursA}\arm{A7} \method{} & \cellcolor{oursA}0.340 & \cellcolor{oursA}\textbf{0.134} & \cellcolor{oursA}0.832 & \multicolumn{3}{c}{\cellcolor{oursA}$+0.097$ \ci{0.030}{0.160} / $-0.034$ \ci{-0.068}{-0.000}} \\
\bottomrule
\end{tabular}}
\caption{Replication of the main arms. Seed 2 repeats the Llama pipeline with a new training seed; the Qwen2.5-7B-Instruct pipeline uses Qwen as its own teacher; the paraphrase set uses author-written phrasings and new tasks. In-distribution: Appendix~\ref{app:more}.}
\label{tab:repl}
\end{table}

Table~\ref{tab:repl} repeats the comparison on a second training seed, a second backbone and unseen phrasings. The arc of Table~\ref{tab:main} reproduces in each: self-training raises persistence and, on Llama, relapse (0.263 on seed 2, 0.268 on the paraphrase set); Stage III cuts relapse relative to \arm{A4} in every setting, by 13--17 points on Llama ($-0.166$ \ci{-0.199}{-0.135} on seed 2; $-0.134$ \ci{-0.166}{-0.104} on paraphrases) and by 7.6 points on Qwen ($-0.076$ \ci{-0.101}{-0.053}), whose backbone relapse is already high and whose SFT stage did not raise it further, and lifts conversation success over the backbone by 10--19 points with intervals excluding zero everywhere. Relapse relative to the backbone is significantly lower on seed 1 and on Qwen, crosses zero on seed 2 (upper bound $+0.004$) and touches it on the paraphrase set; the strict two-part replication gate therefore passes on Qwen and fails narrowly on Llama seed 2. On Qwen, \method{} beats the oracle ledger on both success ($+0.097$ \ci{0.037}{0.157}) and relapse ($-0.100$ \ci{-0.133}{-0.068}); on Llama it beats the ledger on relapse in every setting, matches it on success within the interval on both seeds, and trails it on the paraphrase set (0.340 vs.\ 0.403).

%% file: sec_analysis.tex
\section{Analysis}
\label{sec:analysis}

\subsection{Ablation Studies: What the Preference Stage Needs}
\label{sec:ablation}

\begin{table}[t]
\centering\footnotesize
\setlength{\tabcolsep}{4pt}
\resizebox{\textwidth}{!}{%
\begin{tabular}{llcccccc}
\toprule
& & \multicolumn{3}{c}{\textbf{Held-out}} & \multicolumn{3}{c}{\textbf{In-distribution}} \\
\cmidrule(lr){3-5}\cmidrule(lr){6-8}
\textbf{Arm} & \textbf{Preference pairs} & \textbf{Success} & \textbf{Relapse} & \textbf{Persist.} & \textbf{Success} & \textbf{Relapse} & \textbf{Persist.} \\
\midrule
\arm{A4} (base for DPO) & none & 0.327 & 0.246 & 0.853 & 0.350 & 0.258 & 0.937 \\
\arm{A7}-shuf & \arm{A7}'s 931 pairs, 50\% labels swapped & 0.320 & 0.242 & 0.849 & 0.379 & 0.240 & 0.936 \\
\arm{A7}-act & active-violation pairs only (931) & 0.317 & 0.284 & \textbf{0.881} & 0.412 & 0.308 & \textbf{0.979} \\
\arm{A7}-rel & relapse pairs only (663) & 0.343 & 0.142 & 0.827 & 0.463 & 0.102 & 0.896 \\
\cellcolor{oursA}\arm{A7} \method{} & \cellcolor{oursA}relapse 663 + active 284 & \cellcolor{oursA}\textbf{0.370} & \cellcolor{oursA}\textbf{0.094} & \cellcolor{oursA}0.834 & \cellcolor{oursA}\textbf{0.554} & \cellcolor{oursA}\textbf{0.042} & \cellcolor{oursA}0.904 \\
\bottomrule
\end{tabular}}
\caption{Stage-III ablations (Llama-3.1-8B, same base, hyper-parameters and pair pool as \arm{A7}; preregistered). Pair counts are before length filtering (Appendix~\ref{app:train}).}
\label{tab:ablation}
\end{table}

Table~\ref{tab:ablation} varies only the preference pairs. \emph{Label-shuffled pairs do nothing}: \arm{A7}-shuf is indistinguishable from \arm{A4} on every metric (relapse $-0.004$ \ci{-0.023}{0.014}; success $-0.007$ \ci{-0.043}{0.030}), so the gain is the preference signal and not further training on the distribution. \emph{The relapse-specific contrast is necessary}: pairs that only contrast an active-constraint violation with a passing response raise persistence ($+0.028$ \ci{0.007}{0.051} over \arm{A4}) and \emph{raise} relapse ($+0.038$ \ci{0.015}{0.064}), the same coupling seen in single-turn training (\arm{A1}); \method{} is 19 points below this variant on relapse ($-0.190$ \ci{-0.226}{-0.156}). \emph{Relapse pairs alone recover most of the effect} ($-0.105$ \ci{-0.137}{-0.073} vs.\ \arm{A4}) but the 30\% active-violation pairs add a further 4.7 points of relapse reduction ($-0.047$ \ci{-0.072}{-0.026}) and keep persistence from slipping (\arm{A7}-rel loses 2.6 points of persistence to \arm{A4}; \arm{A7} loses 1.9 with an interval covering zero). Both preregistered ablation gates pass.

\subsection{The Cost of Specialization and a Controllable Trade-Off}
\label{sec:breadth}

\begin{table}[t]
\centering\footnotesize
\setlength{\tabcolsep}{3.5pt}
\resizebox{\textwidth}{!}{%
\begin{tabular}{lcccccccc}
\toprule
& \multicolumn{3}{c}{\textbf{Held-out (ours)}} & \multicolumn{2}{c}{\textbf{IFEval}} & \multicolumn{3}{c}{\textbf{Multi-IF (prompt-strict)}} \\
\cmidrule(lr){2-4}\cmidrule(lr){5-6}\cmidrule(lr){7-9}
\textbf{Model} & \textbf{Success} & \textbf{Relapse} & \textbf{Persist.} & \textbf{strict} & \textbf{loose} & \textbf{T1} & \textbf{T2} & \textbf{T3} \\
\midrule
\arm{A0} backbone & 0.290 & 0.130 & 0.799 & \textbf{0.725} & \textbf{0.771} & \textbf{0.735} & 0.557 & 0.400 \\
\arm{A6b} oracle ledger & 0.390 & 0.148 & 0.877 & -- & -- & -- & -- & -- \\
\arm{A4} self-training & 0.327 & 0.246 & 0.853 & 0.667 & 0.697 & 0.702 & 0.508 & 0.351 \\
\arm{A7} \method{} ($\alpha=1$) & 0.370 & \textbf{0.094} & 0.834 & 0.660 & 0.686 & 0.678 & 0.496 & 0.336 \\
\arm{A7}, state-free inference & -- & -- & -- & 0.636 & 0.689 & -- & -- & -- \\
interpolation $\alpha=0.75$ & 0.397 & 0.125 & 0.857 & 0.686 & 0.723 & 0.695 & 0.516 & 0.374 \\
interpolation $\alpha=0.5$ & \textbf{0.413} & 0.152 & \textbf{0.885} & 0.706 & 0.728 & 0.725 & \textbf{0.560} & \textbf{0.407} \\
\bottomrule
\end{tabular}}
\caption{Breadth. Multi-IF (909 English conversations) accumulates constraints without ever revoking them; IFEval (541 prompts) is single-turn. $\theta_\alpha=\theta_{\text{base}}+\alpha(\theta_{\arm{A7}}-\theta_{\text{base}})$ with no training; both $\alpha$ values were fixed in advance.}
\label{tab:breadth}
\end{table}

Specializing an 8B model to revocable formatting constraints from a 12-family library costs breadth (Table~\ref{tab:breadth}). On Multi-IF, where constraints only accumulate and most of IFEval's 25 instruction types are outside our training families, the trained arms are worse than the backbone at every turn (\arm{A7}: $-6.5$ points at turn 3), and on single-turn IFEval \arm{A7} is $6.5$ points below the backbone on prompt-strict accuracy, failing the preregistered 2-point no-regression gate; a weak pairwise judge on MT-Bench first turns sees a small regression (win 5\%, tie 80\%, loss 15\%). This is a real cost, not a tooling artifact, and we report it as such.

Weight interpolation with the backbone exposes the trade-off as a single knob. At $\alpha=0.5$ the model is within 1.9 points of the backbone on IFEval and, on Multi-IF, within one point of it at turn 1 and above it at turns 2--3 ($+0.7$ at turn 3), while having the highest conversation success (0.413, $+0.123$ \ci{0.057}{0.187} over \arm{A0}) and persistence ($+0.087$ \ci{0.059}{0.114}) of any arm and matching the oracle ledger on every held-out metric (success $+0.023$ \ci{-0.037}{0.083}, relapse $+0.004$ \ci{-0.023}{0.031}). What it gives back is the relapse gain: relapse returns to the backbone level ($+0.022$ \ci{-0.007}{0.051}). The preregistered interpolation gate---IFEval within 2 points \emph{and} relapse below the backbone with an interval excluding zero---is therefore not met by either $\alpha$, and we do not select an $\alpha$ post hoc. The honest summary is a Pareto frontier: $\alpha=1$ for the lowest relapse, $\alpha=0.5$ for the broadest model that still tracks its state, with the untested mitigation being to mix general instruction data into Stages I--III so that the frontier itself moves.

\subsection{Where the Relapse Goes}
Per-turn curves (Appendix~\ref{app:more}) show that the backbone's persistence declines monotonically with depth while relapse is flat at 0.09--0.13 on every post-edit turn; \arm{A4} lifts persistence at all depths and lifts relapse in the same way, and \arm{A7} keeps the persistence curve while pulling relapse below the backbone at every depth. By family, relapse concentrates on constraints that are easy to keep enacting silently once learned---\texttt{no\_commas}, \texttt{no\_digits}, \texttt{sentence\_count}, \texttt{paragraphs\_exact}---and Stage III removes most of it without touching the state block (state exact-match is unchanged), which is the behavioral signature of a model that knew the rule was withdrawn and had not been told that this mattered.

%% file: sec_conclusion.tex
\section{Conclusion}
Evolving-constraint dialogue exposes two failures in instruction-tuned LLMs, forgetting active constraints and re-enacting withdrawn ones, and the only prior repair we are aware of wraps the model in an external ledger. We showed that a SimRAG-style self-improvement recipe with an exact round-trip filter teaches a 7--8B model to track its own constraint state almost perfectly without teaching it to obey that state: on Llama-3.1-8B every supervised variant roughly doubles relapse, and on Qwen2.5-7B none reduces it. Tracking and obeying are separable skills, and a filter can only teach the first. Turning the same verifiable check into a preference signal supplies what the filter cannot, a negative signal against relapse, and 59 steps of DPO on self-labeled contrasts cut relapse by 8--22 points, below the backbone and below an oracle compiled ledger, on two backbones, two seeds and unseen phrasings, with ablations that isolate the relapse-specific contrast as the necessary ingredient. The recipe costs breadth on cumulative-constraint and single-turn benchmarks, and interpolating with the backbone trades that cost against the relapse gain along a single knob. Every number on our evaluation sets comes from executable checkers against compiled states, and training uses no proprietary model, judge or human dialogue annotation; the only judge-based number is the MT-Bench regression signal.

\section{Limitations}
Constraint families are formatting-level and programmatically checkable, and trajectories are scripted from templates with a paraphrase set as the only test of phrasing shift; semantic constraints and user phrasings in the wild are broader. The relapse metric excludes high-chance families by construction and therefore understates relapse on constraints that are easy to satisfy accidentally. Results are for 7--8B backbones with LoRA; stronger models may sit near the floor on relapse, as ReBIND reports, and the breadth cost may differ. The held-out success intervals have half-widths of about five points, and relapse relative to the backbone crosses zero on one seed and touches it on the paraphrase set. The specialization cost on Multi-IF and IFEval is real, the interpolation remedy removes it only by giving back the relapse gain, and the obvious mitigation of mixing general instruction data into training was not run. The MT-Bench judge is a weak local model and its signal is indicative only.

\subsection*{AI use statement}
Large language models were used as writing and coding assistants: to draft and edit prose, to write and refactor training, evaluation and analysis scripts, to format and check references, and to port the manuscript into this template. The hypotheses, preregistered gates, experimental design and interpretation are the authors'; all reported numbers are produced by the released code from result artifacts and were checked by the authors against those artifacts; LLM-generated code was tested before use. We have reviewed all AI-assisted work and take responsibility for the final content of this work, including text, claims and artifacts produced with the aid of generative AI.

\subsection*{Reproducibility statement}
Backbones (Llama-3.1-8B-Instruct, Qwen2.5-7B-Instruct) and benchmarks (Multi-IF, IFEval, MT-Bench) are public. The constraint library, trajectory generator, checkers, state compiler, round-trip filter, pair construction, training configurations and evaluation scripts will be released together with every result JSON (each with a configuration snapshot and content hash), the fixed test-set seeds (Section~\ref{sec:problem}), the preregistered protocol with its amendments (Appendix~\ref{app:prereg}), and the exact hyper-parameters (Appendix~\ref{app:train}). Because Python's \texttt{random} sequences can differ across interpreter versions, the generated trajectories are stored alongside the results.

%% file: sec_statements.tex

%% file: sec_appendix.tex
\section{Constraint Families and Trajectories}
\label{app:families}
Training families (12): \texttt{all\_caps}, \texttt{all\_lower}, \texttt{no\_commas}, \texttt{end\_phrase}, \texttt{start\_phrase}, \texttt{keyword\_count}, \texttt{json\_only}, \texttt{bullets\_exact}, \texttt{max\_words}, \texttt{forbidden\_word}, \texttt{paragraphs\_exact}, \texttt{wrap\_quotes}. Held-out families (6, test only): \texttt{no\_digits}, \texttt{sentence\_count}, \texttt{title\_brackets}, \texttt{postscript}, \texttt{two\_responses}, \texttt{numbered\_sections}. Each family has an executable checker and one to three natural-language phrasings per operation (\textsc{add}, \textsc{amend}, \textsc{revoke}); the paraphrase set uses three further author-written phrasings per family and operation and new task prompts. Measured chance-satisfaction rates $p_0$ on the backbone's unconstrained responses are below 0.05 for every family except \texttt{forbidden\_word} (0.94--0.95), which is therefore excluded from relapse, and \texttt{paragraphs\_exact} (0.04) and \texttt{bullets\_exact} (0.02), which are retained. Trajectories are generated from a seed with 4--8 turns, at most three active constraints, incompatible pairs never jointly active, a \textsc{distract}/\textsc{return} pair in half of the trajectories, and at least one \textsc{revoke} or \textsc{amend} per trajectory. Test sets: held-out seed 777 (300 conversations), in-distribution seed 20260916 (240), paraphrase seed 999 (300); training trajectories use disjoint seeds.

\section{Training Details and Filter Statistics}
\label{app:train}
\paragraph{SFT stages.} LoRA rank 64, $\alpha=128$, dropout 0.05 on all attention and MLP projections; learning rate $10^{-4}$, 3\% warm-up, one epoch, effective batch 16 sequences, maximum length 4,096; identical for every SFT arm, only the data differs. Stage I (Llama): 9,238 examples, 578 steps, 55--61 minutes on two A100s (loss 0.29 $\rightarrow$ 0.07 for the state arm). Stage II (state arm): 9,920 kept turns, 620 steps, 68 minutes. Single-turn arm \arm{A1}: 2,237 examples, 140 steps.
\paragraph{Stage III (DPO).} Base and frozen reference: the Stage-II state model; LoRA as above; learning rate $2\times10^{-5}$, 5\% warm-up, one epoch, effective batch 16 pairs, $\beta=0.1$, NLL weight 0.5 on the chosen response, at most 30\% active-violation pairs. 931 pairs encoded (663 relapse, 284 active-violation; 16 dropped for length), 59 steps, 17 minutes; training preference accuracy 0.50 $\rightarrow$ 1.00 and loss 0.77 $\rightarrow$ 0.31. Ablations: relapse-only 649 pairs / 41 steps; active-only 922 pairs / 58 steps; label-shuffled 931 pairs / 59 steps. All values were fixed in the protocol amendment of 2026-09-16 16:50 and never tuned.
\paragraph{Filter yields.} A turn is rejected when none of its $K=4$ sampled candidates passes, which also stops the trajectory; a rejected turn is counted under every distinct reason among its candidates, so reason counts slightly exceed rejected-turn counts. Stage-I data generation (backbone as teacher, 4 shards): 10,723 turns attempted, 9,238 kept (yield 0.86); stopping reasons 485 active-constraint violations vs.\ 1,027 relapses. Stage-II generation with the state model: 11,243 attempted, 9,920 kept (0.88); reasons 411 active violations, 887 relapses, 158 state mismatches. Relapse is the dominant rejection reason at every stage and on both backbones (Qwen Stage I: 1,253 relapse vs.\ 446 active-violation stops), which is the data-side signature of the result in Section~\ref{sec:main}.
\label{app:filter}

\section{Preregistration Summary}
\label{app:prereg}
The protocol fixed, before any training run, the arms of Table~\ref{tab:main}, the metrics, the test seeds, and the gates: G0.1 (backbone final-turn persistence $\le 0.80$; pass, 0.653), G0.2 (prompted state exact-match $\ge 0.5$; fail, 0.003), G0.3 (state prompt does not hurt; fail), G1 (\arm{A4} above \arm{A0} on success and below on relapse, intervals excluding zero; fail), G2 (\arm{A4} below \arm{A3} on relapse; narrow pass), H2 (state-free \arm{A5} retains half of the \arm{A4} gain; fail), H3 (\arm{A4} within the interval of \arm{A6b}; fail). Amendment of 2026-09-16 16:50 (before the H5 runs): \arm{A9} and \arm{A7}, gate H5 (pass for \arm{A7}, fail for \arm{A9}). Amendment of 2026-09-17 22:50: seed 2 and Qwen replication (direction replicates; strict two-part gate passes on Qwen, fails narrowly on Llama seed 2), Multi-IF direction (reversed), IFEval no-regression within 2 points (fail, $-6.5$), MT-Bench loss-minus-win within 10 points (boundary; treated as fail), paraphrase relapse below backbone with interval excluding zero (upper bound $-0.000$; marginal miss). Amendment of 2026-09-18 22:50: Stage-III ablations (both gates pass) and interpolation (gate fails; both $\alpha$ reported). No gate or hyper-parameter was changed after results were seen.

\section{Additional Results}
\label{app:more}
\paragraph{In-distribution replication.} Llama seed 2 (240 conversations): \arm{A0} success 0.321 / relapse 0.092; \arm{A6b} 0.450 / 0.101; \arm{A4} 0.379 / 0.264; \arm{A7} 0.475 / 0.074. Qwen2.5-7B: \arm{A0} 0.133 / 0.254; \arm{A6b} 0.254 / 0.275; \arm{A4} 0.279 / 0.284; \arm{A7} 0.358 / 0.179. Stage-III ablations in distribution (Table~\ref{tab:ablation}): \arm{A7} vs.\ \arm{A7}-act relapse $-0.266$ \ci{-0.309}{-0.224}; vs.\ \arm{A7}-shuf $-0.199$ \ci{-0.240}{-0.161}; vs.\ \arm{A7}-rel $-0.061$ \ci{-0.090}{-0.033} with success $+0.092$ \ci{0.042}{0.146}.
\paragraph{Per-turn behavior.} Backbone persistence by turn index on the diagnostic set: 0.887, 0.842, 0.803, 0.771, 0.741, 0.712, 0.681, 0.653; relapse on post-edit turns 0.09--0.13 with no trend. Trained arms shift the whole persistence curve upward by 4--9 points; \arm{A7} lowers relapse below the backbone at every depth, \arm{A4} raises it at every depth. Per-family relapse counts for \arm{A4} concentrate on \texttt{no\_digits}, \texttt{sentence\_count}, \texttt{no\_commas} and \texttt{paragraphs\_exact}; Stage III reduces each by more than half.
\paragraph{Interpolation vs.\ oracle ledger.} $\alpha=0.5$ vs.\ \arm{A6b} (held-out, paired): success $+0.023$ \ci{-0.037}{0.083}, relapse $+0.004$ \ci{-0.023}{0.031}, persistence $+0.008$ \ci{-0.015}{0.034}. $\alpha=0.75$ vs.\ \arm{A0}: success $+0.107$ \ci{0.047}{0.170}, relapse $-0.005$ \ci{-0.038}{0.027}, persistence $+0.058$ \ci{0.026}{0.089}; IFEval prompt-strict 0.686, Multi-IF turn 3 0.374.
\paragraph{Prompts.} The state-mode system prompt asks the model to begin its reply with a fenced JSON block containing \texttt{active} (list of \{family, params\}) and \texttt{revoked} (same schema) and then to answer; the block is stripped before checking. State-free mode uses the backbone's default system prompt. The ledger baselines prepend the compiled specification (\arm{A6b}: rules in force; \arm{A6}: rules in force plus a ``do not follow'' list of withdrawn rules) to every user turn, matched in token count to the state block.